\pdfoutput=1

\documentclass[11pt]{article}

\usepackage{ACL2023}

\usepackage{times}
\usepackage{latexsym}

\usepackage[T1]{fontenc}

\usepackage[utf8]{inputenc}

\usepackage{microtype}

\usepackage{inconsolata}
\usepackage{array}
\usepackage{amssymb}
\usepackage{graphicx}
\usepackage{multirow}
\usepackage{booktabs,arydshln}

\makeatletter
\def\adl@drawiv#1#2#3{%
        \hskip.5\tabcolsep
        \xleaders#3{#2.5\@tempdimb #1{1}#2.5\@tempdimb}%
                #2\z@ plus1fil minus1fil\relax
        \hskip.5\tabcolsep}
\newcommand{\cdashlinelr}[1]{%
  \noalign{\vskip\aboverulesep
           \global\let\@dashdrawstore\adl@draw
           \global\let\adl@draw\adl@drawiv}
  \cdashline{#1}
  \noalign{\global\let\adl@draw\@dashdrawstore
           \vskip\belowrulesep}}
\makeatother

\title{Extracting Text Representations for Terms and Phrases\\in Technical Domains}

\author{Francesco Fusco$^*$\\
IBM Research \\
\texttt{ffu@zurich.ibm.com} \\\And
Diego Antognini$^*$\\
IBM Research\\
\texttt{Diego.Antognini@ibm.com} \\}

\begin{document}
\maketitle
\def\thefootnote{*}\footnotetext{Equal contribution.}\def\thefootnote{\arabic{footnote}}

\begin{abstract}

    Extracting dense representations for terms and phrases is a task of great
    importance for knowledge discovery platforms targeting highly-technical
    fields. Dense representations are used as features for downstream components
    and have multiple applications ranging from ranking results in search to
    summarization. Common approaches to create dense representations include
    training domain-specific embeddings with self-supervised setups or using
    sentence encoder models trained over similarity tasks. In contrast to static
    embeddings, sentence encoders
    do not suffer from the out-of-vocabulary~(OOV) problem, but impose significant
    computational costs. In this paper, we propose a fully \textit{unsupervised approach}
    to text encoding that consists of training small character-based models with the
    objective of \textit{reconstructing} large pre-trained embedding matrices.
    Models trained with this approach can not only match the quality of sentence encoders
    in technical domains, but are \textit{5 times} smaller and up to \textit{10~times}
    faster, even on high-end~GPUs.

\end{abstract}

\section{Introduction}

Large pre-trained language models are extensively used in modern NLP systems.
While the most typical application of language models is fine-tuning to
specific downstream tasks, language models are often used as text encoders to
create dense features consumed by downstream components. Among the many use
cases of dense text representations there is search, question answering, and
classification~\cite{yang-etal-2020-multilingual}.

Static embeddings, trained with algorithms such as
Word2Vec~\cite{word2vec}, can exploit existing information extraction pipelines to
create representations for entities, phrases, and terms present in text
corpora. Static embedding matrices are
trained with self-supervised approaches at regular intervals, either when additional
data is available or to leverage improvements in information extraction models.
Pre-trained embedding matrices can be considered as static feature stores, providing
dense representations for entries belonging to a fixed vocabulary. Representations
for entries outside of the vocabulary are not available, leading to the out-of-vocabulary~(OOV) problem.

In contrast, contextualized word embeddings leverage sentence encoders~\cite{cer-etal-2018-universal,reimers-gurevych-2019-sentence} 
to dynamically create dense
representations for any input text by performing a forward pass over a large
language model. Specifically, a word embedding is computed at inference time 
based on its context, unlike static word embeddings that have a fixed (context-independent) 
representation. In practice, sentence encoders solve the
out-of-vocabulary~(OOV) problem which affects static embeddings at the cost of
high computational requirements and stronger dependencies on supervised
datasets for similarity.

Despite the popularity of sentence encoders, large pre-trained embedding
matrices are still widely adopted in the industry to encode not only individual
tokens but also multi-token entities extracted with in-house NLP pipelines.
Once those embedding matrices are trained, the text representation for single-
and multi-token entries encountered at \textit{training time} can be looked up
in constant time.

In this paper, we describe an effective approach taken to provide
high-quality textual representations for terms and phrases in a commercially
available platform targeting highly-technical domains. Our
contribution is a novel \textit{unsupervised approach} to train text encoders
that bridges the gap between large pre-trained embedding matrices
and computationally expensive sentence encoders. In a nutshell, we exploit the
vast knowledge encoded in large embedding matrices to train small character-based
models with the objective of \textit{reconstructing} them, i.e., we use large
embedding matrices trained with self-supervision as large \textit{training
datasets} mapping text to the embedding vectors.

Our approach is extremely attractive for industrial setups as it 
leverages continuous improvements, testing, and inference costs
of existing information extraction pipelines to create large datasets to train
text encoders. This way, the return on investment for annotations, development,
and infrastructure costs are maximized. 

In our evaluation, we highlight that by combining unsupervised term extraction
annotators and static embeddings we can train lightweight character-based
models that \textit{match} the quality of supervised sentence encoders and
provide \textit{substantially better} representations than sentence encoders
trained without supervision. Our models not only provide competitive
representations, but are up to \textit{5 times} smaller and \textit{10 times}
faster than sentence encoders based on large language models.

\section{Existing Approaches}

The mission of our industrial knowledge discovery platform is to extract
knowledge from large corpora
containing highly-technical documents, such as
patents and papers, from diverse fields ranging from chemistry, to physics,
to computer science. Information extraction is extremely challenging given the
large variety of language nuances and the large cost of human annotations in such
specialized domains. Therefore, it is of extreme importance to minimize
the dependencies on annotated data and to use unsupervised
approaches whenever possible.

A recurring requirement from many internal components of our platform is the
ability to extract high-quality dense representations for technical terms,
entities, or phrases which can be encountered in many distinct technical fields.
High-quality representations are extremely valuable to implement semantic
search, to influence the ranking, or to be used directly as model features.

In modern industrial systems, it is often the case that static 
and context-dependent embedding technologies coexist on the same platform to
extract representations.
While static embeddings are trained in a self-supervised fashion, sentence encoders 
are often built by fine-tuning pre-trained models on similarity tasks using
annotated datasets. Having two separate approaches for text encoding is
suboptimal as those systems are \textit{completely independent} and embed terms
into \textit{distinct embedding spaces}.

To reconcile those two worlds, we propose an approach where static embeddings
and text encoders are mapping text into the \textit{same embedding space}. Our intuition
is that static embedding matrices storing embeddings for single tokens, but
also multi-token terms, entities, or phrases, represent an \textit{invaluable source of
information to train text encoders}. While those matrices are built with
self-supervision, they can leverage existing annotators, supervised or not,
which are commonly available in large text processing platforms.

Our novel approach consists of using pre-trained embedding matrices as a
training set, and training character-based models, called CharEmb, to predict the embedding
vector for a text. This means that the character-based models will enable to project
\textit{any} sequence of characters in the same embedding space as the pre-trained embedding
matrices.

\begin{figure*}[ht!]                                                                    
\centering                                                                              
\includegraphics[width=.92\textwidth]{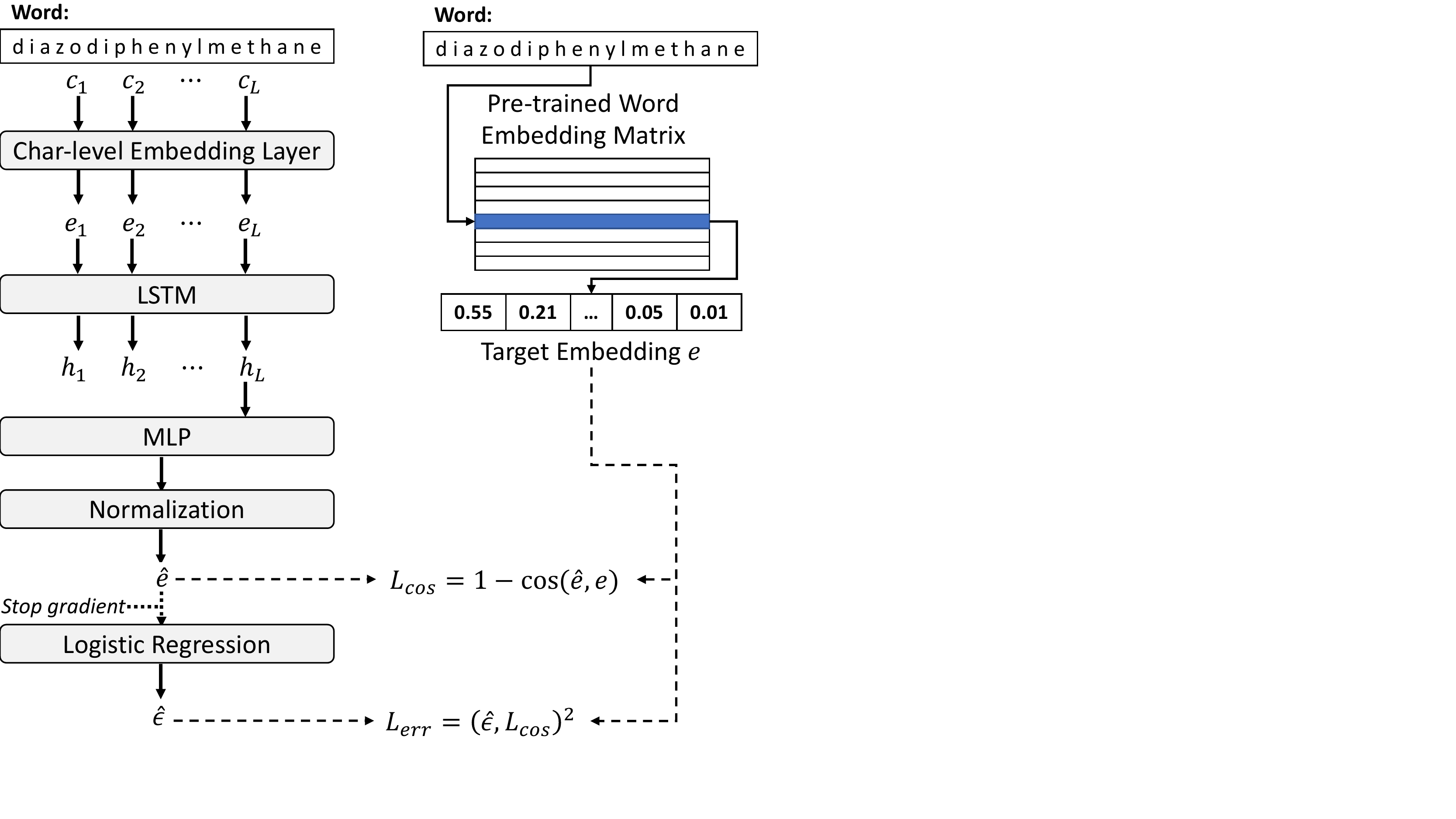}                                                                                                                                       
\caption{Illustration of the data flow for our approach. We introduce CharEmb, a character-based model
    that projects a given text into the same embedding space of a large pre-trained embedding model.
    CharEmb uses the pre-trained embedding only at training to learn the projection function between a text and the its embedding vector.}
\label{fig:workflow}                                                                    
\end{figure*}

\subsection{Static Embeddings}

Algorithms to train static word embeddings, such as Word2Vec~\cite{word2vec}, have been
introduced to efficiently compute the representations for words or entire
phrases extracted from large text corpora. To create the representation of
entire phrases, the original Word2Vec paper suggested
to simply preprocess the text to make sure that phrases are treated
as distinct words in the vocabulary. 
In a nutshell, the preprocessing step involves merging
the tokens of which a phrase is composed into a unit which is not split by the
tokenizer, e.g., \texttt{[}\textit{``machine''},
 \textit{``learning''}\texttt{]} becomes
\texttt{[}\textit{``machine\_learning''}\texttt{]}.

In a typical industrial setup, this approach can be naturally generalized to
leverage the large set of annotators that are commonly available in
large-scale natural language processing platforms. This way one can create
domain-specific embeddings not just for single tokens, but for entities, terms, phrases which
are extracted in a natural language processing platform. Combining
self-supervised word embedding algorithms together with existing sequence
tagging models is extremely attractive. First, one can fully leverage the
constant enhancements of in-house models to improve the
quality of the embeddings for all the entities of interests. Second, since the
sequence tagging models are built in-house and independently evaluated, using
them to build embedding matrices means reducing the time spent in quality
assurance~(QA). Third, since the model inference is anyway computed over large
amount of textual data while the natural language processing platform is
operating, one can amortize that compute costs to accelerate another task,
i.e., providing high-quality text representation.

\subsection{Sentence Embeddings}

Static embedding matrices built with the previous approach can provide
representations for several hundred millions entries when trained over large
text corpora pre-processed with several name entity recognition~(NER) models.
Despite the large size, one can still experience the problem of
out-of-vocabulary~(OOV), which means, downstream components might require text
representations for text entries which are not present in the vocabulary.

Text encoders have been introduced to solve the OOV problem. They provide ways
to create embeddings that are not static but contextualized, which means that
the embedding vector must be created on the fly via a model inference.
Contextualized embeddings can be created using pre-trained models trained with
self-supervised setups such as BERT~\cite{devlin-etal-2019-bert} or with text encoders which are still based
on large pre-trained models, but fine-tuned with task similarity datasets.
Sentence encoders trained with supervised setups using for example the NLI
datasets~\cite{bowman-etal-2015-large,williams-etal-2018-broad}, such as S-BERT~\cite{reimers-gurevych-2019-sentence} are well known to perform well in practice to create representations for entire sentences or
features for finer grained text snippets. The limitation of supervised
approaches for sentence encoding is that \textit{creating large annotated datasets for
similarity is extremely expensive}, especially in technical fields. Therefore,
improving the sentence encoder themself requires substantial investments in
annotations. Unsupervised sentence encoder
approaches~\cite{gao-etal-2021-simcse,wang-etal-2021-tsdae-using}, on the other
hand are well known to offer poorer performance than supervised counterparts.

\section{Our Model: CharEmb}
Instead of having two completely disjoint systems to create text
representations, we use character-based models trained over large static
embedding matrices, that project a sequence of text into the same
embedding space as the embedding matrices used as training data. In practice,
\textit{we approach text encoding as a compression problem}, where character-based
models are trained to reconstruct the pre-trained static embedding matrices, as shown
in Figure~\ref{fig:workflow}.
This training approach can rely on supervised sequence
tagging models or can be implemented using fully unsupervised
methods, such as the term extraction technologies described in the
literature~\cite{termExtraction2022}. As we highlight in the evaluation
section, a text encoder trained without any supervision can match the
performance of supervised sentence encoders in creating representations for
words or phrases in technical domains.

\begin{figure}[t!]                                                                     
    \centering                                                                          
    \includegraphics[width=1.0\linewidth]{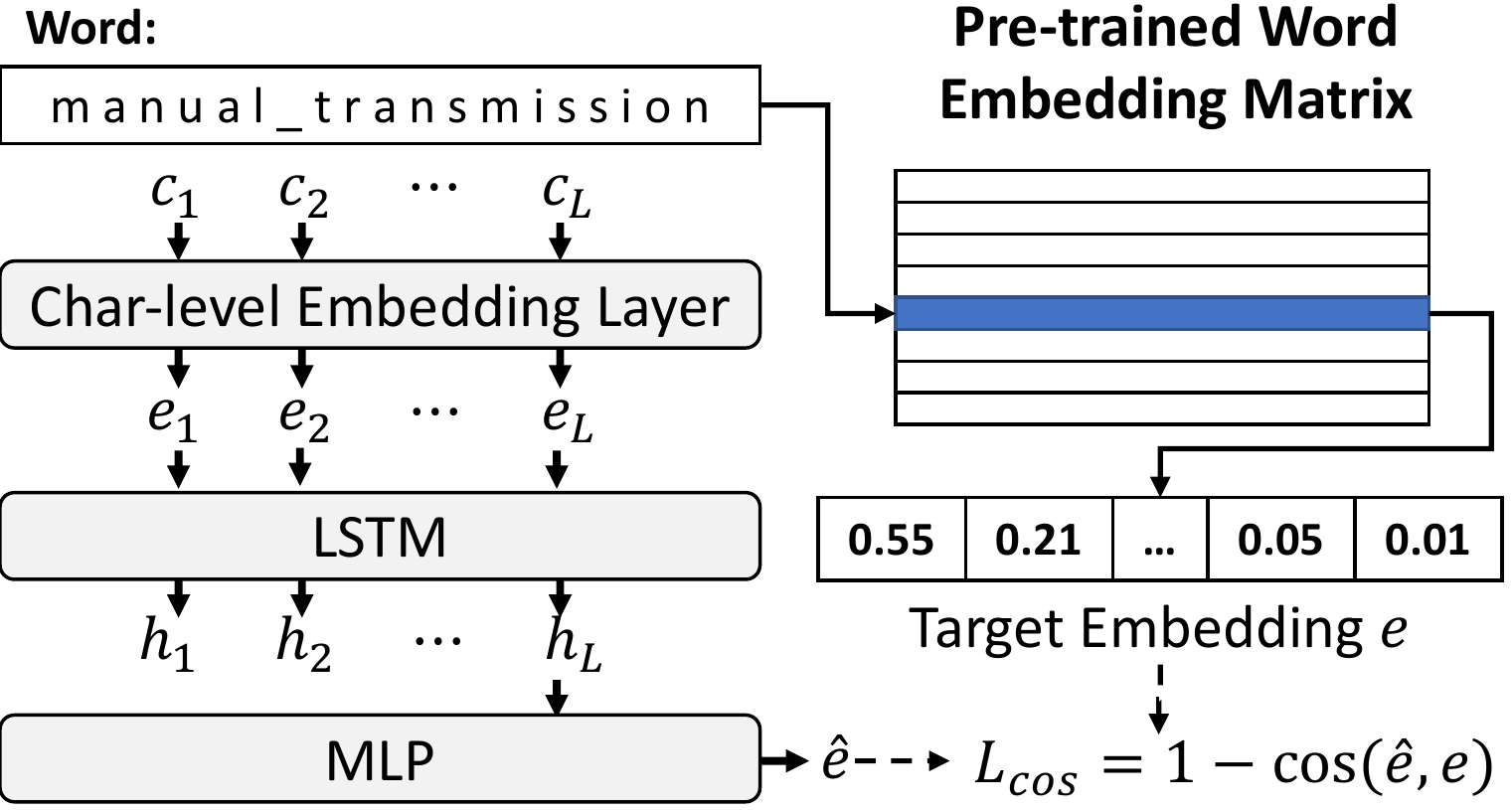}                 
    \caption{CharEmb is trained to predict an embedding that has a close cosine similarity with the target~one.}                         
    \label{fig:model} 

\end{figure}

To train our models, we consider a static pre-trained embedding matrix as gold
dataset. An individual training sample associates a single- or multi-token text
to an embedding vector. To leverage the
dataset we train a text encoder model to minimize the cosine similarity
between the produced vector and the original vector stored in the static
embedding matrix. The models rely on character-level tokenization to generalize
better on unseen inputs in technical domains.
Figure~\ref{fig:model} highlights a simple yet effective LSTM-based model
architecture. The pre-trained static embedding matrix~(on the right) contains
$|V|$ embedding vectors of size $k$, where $V$ is the vocabulary of the static embedding
matrix. The model receives as
input the text $t$, tokenize it in characters, for which an embedding matrix is
trained. Character-level embeddings are used as input for a Long Short-Term
Memory sequence
model. The last state of the LSTM layer is used to produce, via a Multi-Layer Perceptron, a
vector of dimension $k$ that represents the
embedding for the text $t$. The network is trained to minimize the cosine
distance between the predicted embedding and the original one
stored in the embedding matrix. The number of distinct training samples
is $|V|$, which means that embeddings with large vocabularies correspond to bigger
training datasets.

\section{Evaluation}
\label{sec:evaluation}

In this section, we compare our text representations using the \textit{Patent Phrase Similarity
Dataset} built by Google \cite{https://doi.org/10.48550/arxiv.2208.01171}.
Given two multiword technical concepts (called \textit{anchor} and \textit{target}), the task consists of predicting a similarity score between the two terms. Unlike general-purpose sentence datasets, such as STS-B~\cite{cer-etal-2017-semeval} or SICK~\cite{marelli-etal-2014-sick}, we focus on technical concepts in the patent and scientific domains. The dataset contains $38,771$~unique concepts and $36,473$, $2,843$, and $9,232$ concept pairs with humanly-annotated similarity scores for the train, validation, and test sets, respectively. 

In our case, we are only interested in the \textit{zero-shot} scenario, and thus, we \textit{only} consider the test set and ignore the train and validation sets. We evaluate the quality of the text representations using the same approach described in~\citet{https://doi.org/10.48550/arxiv.2208.01171}: we compute the Pearson and~Spearman correlation between the cosine similarity of the two embeddings and the human-annotated scores.\footnote{For reproducibility purposes, we include all experimental details and the hyperparameters in Appendix~\ref{app:training}.}

\subsection{Static Pre-trained Word Embeddings}
\label{sec:static}

First, we compare the performance of publicly-available pre-trained embedding
matrices with embeddings trained with in-domain data. Following the
approach described in \citet{https://doi.org/10.48550/arxiv.2208.01171}, we
compute the representation for concepts consisting of multiple tokens as the
average of the embedding vectors of each unigram.

To highlight the importance of in domain-data, we train static embedding
matrices using a relatively small corpus consisting of 120 million
sentences sampled from the the ArXiv~\cite{clement2019arxiv} and the
HUPD~\cite{suzgun2022hupd} datasets. To train our embeddings, we pre-process
the text after running term-extraction using the unsupervised method described
in \citet{termExtraction2022}. This way, our method can be considered fully
unsupervised, and its evaluation does not depend on proprietary
annotations and model architectures.

The size of the text after pre-processing is 18~Gigabytes accounting for 1.9
billion tokens with term-extraction enabled and 2.2 billion tokens without. 
We train the static embeddings using~CBOW \cite{word2vec}. Training for one 
epoch~takes 25 minutes on a 16-core 
\textit{AMD EPYC 7742}, which corresponds to less than \textit{10 dollars} of
compute costs with current cloud offerings.
We do not consider the term extraction as part of the
overall training costs since in practice, the large amount of annotations that
a large-scale NLP platform produces during its execution can be entirely~reused.

Finally, we train three variants.~The~first contains unigrams and multiword
expressions extracted with our term extractor~represented with one token
(i.e., ``machine learning'' $\rightarrow$ ``machine\_learning''). The second
considers only unigrams (i.e., with the term-extraction component disabled).
For the third we use FastText \cite{bojanowski2016enriching} instead.

We compare our embedding matrices with the official pre-trained models for
GloVe~\cite{Glove}, Word2Vec~\cite{bojanowski2016enriching}, and
FastText~\cite{bojanowski2016enriching}. Those are trained on corpora
that are substantially larger than our ArXiv-HUPD dataset~(up to \textit{300~times}).

Table~\ref{tab:context_indep_emb} reports the Pearson and Spearman correlations
when using the representations of the static word embedding matrices. Not
surprisingly the embeddings trained over the ArXiv-HUPD corpus, which contains
text of the same domain, provide substantially better results than embeddings
pre-trained over corpora that are out-of-domain, such as news and crawled
websites. Our embeddings trained on only 2 billion tokens outperform embeddings trained
over corpora that are up to \textit{2 order of magnitude larger}.
Further, we see that our static-embedding matrices including
terms are providing only a marginal improvement, as the terms do not necessarily
cover concepts present in the~dataset.

\begin{table}[!t]
\small
\centering
    \begin{tabular}{
    @{}
    l@{\hspace{.5mm}}
    c@{\hspace{1mm}}
    c@{\hspace{1mm}}
    c@{\hspace{1mm}}
    c@{\hspace{1mm}}
    c@{}}
         & & & & \multicolumn{2}{c}{Correlation}\\
         \cmidrule{5-6}
        Pre-trained embedding & |Voc.| & Size (MB) & Dim. & Pear. & Spear.\\
\toprule
GloVe (6B) & 0.4M & 458 & 300 & $42.37$ & $43.95$\\
GloVe (42B) & 1.9M & 2,194& 300 & $40.30$ & $45.83$\\
GloVe (840B) & 2.2M & 2,513& 300 & $44.83$ & $49.71$\\
FastText wiki-news (16B)& \multirow{1}{*}{1.0M} & \multirow{1}{*}{1,144} & \multirow{1}{*}{300} & $39.01$ & $46.03$\\
FastText crawl (600B)& \multirow{1}{*}{2.0M} & \multirow{1}{*}{2,289} & \multirow{1}{*}{300} & $47.36$ & $49.32$\\
Word2Vec news (100B) & 3.0M & 2,861 & 250 & $44.04$ & $44.72$\\
\cdashlinelr{1-6}
ArXiv-HUPD (Ours) & & & & \\
- uni (FastText) (2.2B) & \multirow{1}{*}{5.3M} & \multirow{1}{*}{6,006} & 300 & $51.25$ & $49.92$\\
- uni (Word2Vec) (2.2B)& \multirow{1}{*}{1.8M} & \multirow{1}{*}{1,403} & 200 & $50.82$ & $52.97$\\
- uni + terms (1.8B)& 5.2M & 3,984 & 200 & $\textbf{51.62}$ & $\textbf{53.91}$\\
    \end{tabular}
    \caption{\label{tab:context_indep_emb}\textbf{Static} context-independent word embeddings. Brackets denote the number of token of the corpus. Training static embeddings on ArXiv-HUPD improves significantly the correlation with the human annotations.}
\end{table}
\begin{table}[!t]
\small
\centering
    \begin{tabular}{
    @{}
    l@{\hspace{2mm}}
    c@{\hspace{2mm}}
    c@{}
    c@{\hspace{2mm}}
    c@{}}
         & \multicolumn{3}{c}{Pearson Correlation} &\\
         \cmidrule{2-4}
    Models & Original & Reconstr. & $\Delta$ & Compression\\
\toprule
CharEmb Small {\textsubscript{13MB}} & \multirow{3}{*}{$51.62$} & $49.70$ & $-3.7\%$ & 306x\\
CharEmb Base {\textsubscript{38MB}} & & $54.33$ & $+5.3\%$ & 236x\\
CharEmb Large {\textsubscript{86MB}} & & $55.97$ & $+8.4\%$ & \ \ 46x\\
    \end{tabular}
    \caption{\label{tab:compression}\textbf{Reconstructed static} word embeddings. We report the correlation of the original ArXiv-HUPD embeddings (uni + terms) and the reconstructed ones inferred by our models. CharEmb Base achieves a compression by a factor of 236 and an improvement of $+5.3\%$.}
\end{table}

After focusing on the raw embedding matrices, we evaluate the quality of our
CharEmb models as compressors. In practice, we repeat the same experiments when
the best ArXiv-HUPD word embedding matrix is \textbf{\textit{fully reconstructed}} by
projecting each vocabulary entry using a character-based model trained with our
approach. We report correlations for models based on the Long Short-Term
Memory~\cite{10.1162/neco.1997.9.8.1735}, because in our
experiments, it offered significantly better results than Gated Recurrent
Unit~\cite{69e088c8129341ac89810907fe6b1bfe} and
Transformer~\cite{NIPS2017_3f5ee243}. 
We report the performance for three model sizes:
\textit{Small} (13MB), \textit{Base} (38MB), and \textit{Large} (86MB).

Table~\ref{tab:compression} shows that our base (38MB) and large (86MB) models
compress the embedding matrix they are trained on \textit{and improve its
quality} according to the Pearson correlation (similar trend with Spearman).
This means that a model of solely 38 MB not only can fully reconstruct the
\textit{3.98 GB} matrix it has been trained on, given only its vocabulary~(236x
space reduction), but also that the \textit{reconstructed} matrix provides a
correlation gain of \textit{$+5.3\%$} compared to the original one.

\subsection{Sentence (Contextualized) Embeddings}
\label{sec:context}

\begin{table}[!t]
\small
\centering
    \begin{tabular}{
    @{}
    l@{}
    c@{\hspace{2mm}}
    c@{\hspace{4mm}}
    c@{\hspace{2mm}}
    c@{}}
         & & & \multicolumn{2}{c}{Correlation}\\
         \cmidrule{4-5}
        Models & Size (MB) & Dim. & Pear. & Spear.\\
\toprule
BERT (Unsup.) & \multirow{1}{*}{1,344} & \multirow{1}{*}{1,024} & $43.07$ & $41.40$\\
Patent-BERT (Unsup.) & 1,344 & 1,024 & $54.00$ & $54.47$\\
SimCSE (Unsup.) & \multirow{1}{*}{438} & \multirow{1}{*}{768} & $53.35$	& $51.91$ \\
Patent-SimCSE (Unsup.) & 438 & 768 & $50.51$ & $48.33$\\
Sentence-BERT (Sup.) & 438 & 768 & $59.82$ & $57.66$\\
SimCSE  (Sup.) & 438 & 768 & $56.63$	& $56.81$ \\
\cdashlinelr{1-5}
\multicolumn{5}{c}{\textit{ArXiv-HUPD - unigrams (Word2Vec) (Ours)}}\\
CharEmb Small (Unsup.) & 13 & \multirow{3}{*}{200} & $41.68$ & $39.55$\\
CharEmb Base (Unsup.) & 38 & & $47.57$ & $46.16$\\
CharEmb Large (Unsup.) & 86 & & $47.58$ & $45.60$\\	
\multicolumn{5}{c}{\textit{ArXiv-HUPD - unigrams + terms (Ours)}}\\
CharEmb Small (Unsup.) & 13 & \multirow{3}{*}{200} & $55.53$ & $56.73$\\
CharEmb Base (Unsup.) & 38 & & $58.53$ & $59.66$\\
CharEmb Large (Unsup.) & 86 & & $\textbf{59.84}$& $\textbf{60.52}$\\	

    \end{tabular}
    \caption{\label{tab:context_dep_emb}\textbf{Sentence}-based word embeddings via predictions. CharEmb is unsupervised, at least 5x smaller, and outperforms large supervised and unsupervised baselines. Training CharEmb on ArXiv-HUPD that contains unigrams clearly underperforms, showing how important multiword expressions are during training.}
\end{table}

In Table~\ref{tab:compression} we use our models to reconstruct an original
embedding matrix. The reconstructed matrix is used as a static
pre-trained embedding matrix: given a phrase in the test of the patent dataset,
we compute the representation as the average of the unigrams. Instead, in this
section we use our models as \textit{text encoders}, which means performing
the inference with our CharEmb models 
to extract representations of the phrases present in the test set.

Following \citet{https://doi.org/10.48550/arxiv.2208.01171}, we compare
our CharEmb variants with the following pre-trained models used as text
encoders: BERT~\cite{devlin-etal-2019-bert}, Patent-BERT, and the sentence
encoder Sentence-BERT~\cite{reimers-gurevych-2019-sentence} trained on the
natural language inference
datasets~\cite{bowman-etal-2015-large,williams-etal-2018-broad}. We augment the
proposed baselines with a popular sentence-encoding method
SimCSE~\cite{gao-etal-2021-simcse}, which can be trained with supervision
(similarly to S-BERT) or in an unsupervised manner. For the latter, we include
the publicly-available variant trained on Wikipedia~(Unsupervised SimCSE) and train our own model over
the small ArXiv-HUPD dataset~(Patent-SimCSE).\footnote{Additional results when training CharEmb on
GloVe, FastText, and Word2Vec embeddings are shown in
Appendix~\ref{app:context}.}

In Table~\ref{tab:context_dep_emb}, we report the Pearson and Spearman
correlation when using text encoders to produce text representations via
inferencing to a model. Thanks to our approach, lightweight LSTM-based models
outperform larger BERT-based models in a \textit{zero-shot} setup. Our large
model is \textit{5x smaller} than Sentence-BERT and SimCSE and yet provides
\textit{better representations}. Training CharEmb does not require any manual
annotation, since embeddings are trained with self-supervision and the term
extraction is fully unsupervised. \textit{Our smallest model outperforms all
unsupervised approaches.}
Furthermore, we note that term extraction is a \textit{fundamental component}
for the creation of high-quality CharEmb models. When training over
unigram-only embeddings, our models performance drops significantly to the
levels of BERT.

\begin{figure}[!t]
	\centering
    \includegraphics[width=0.49\textwidth,height=3.05cm]{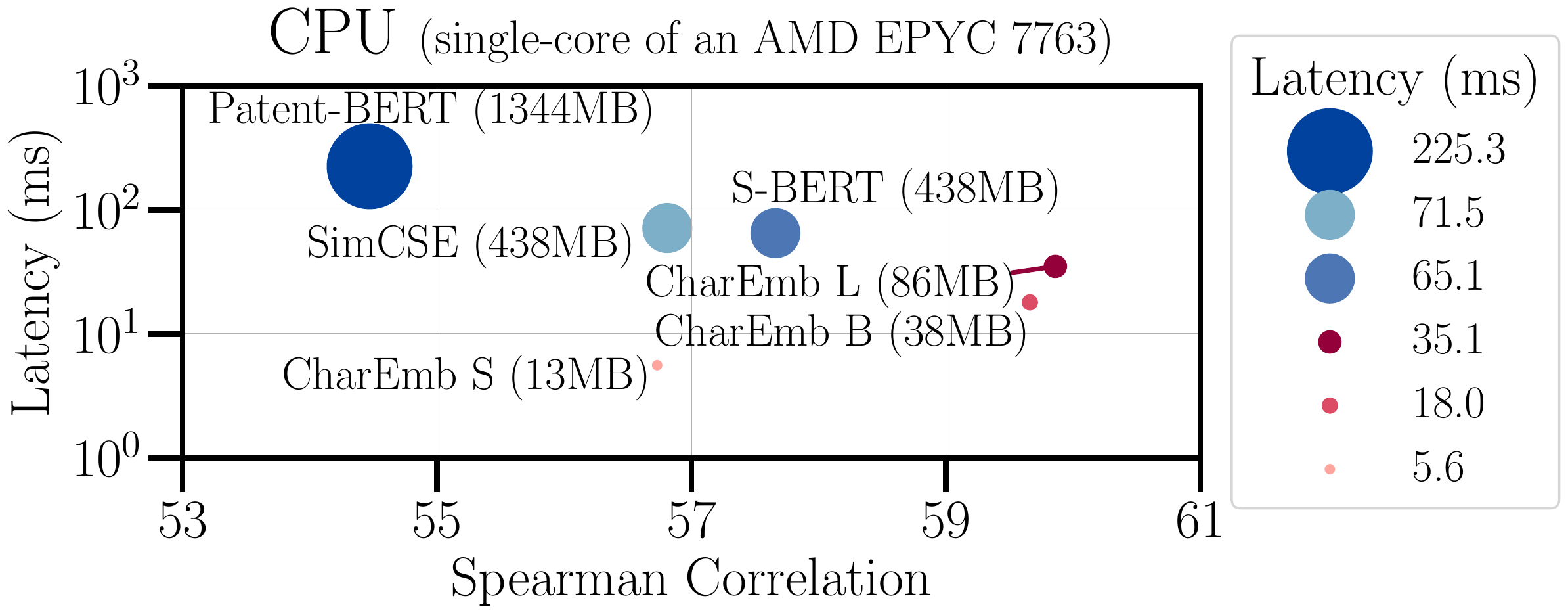}\\
    \includegraphics[width=0.49\textwidth,height=3.05cm]{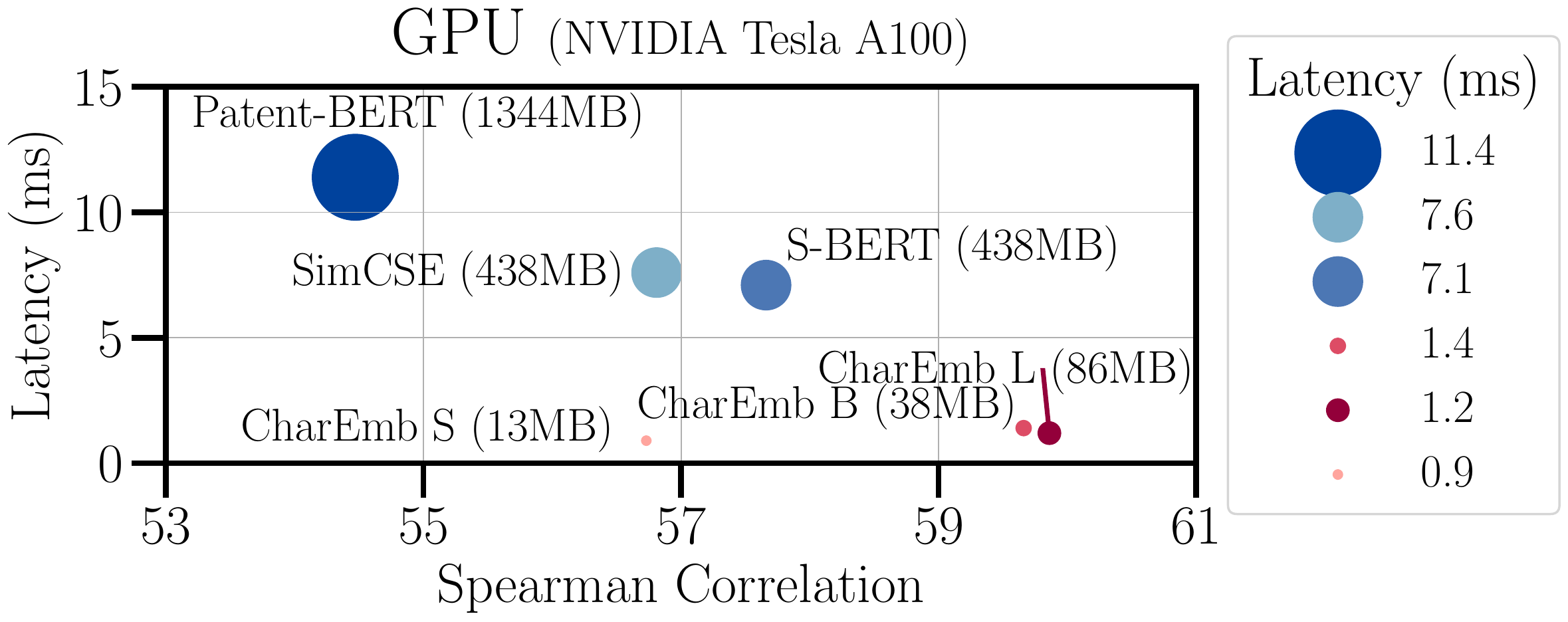}

    \caption{Inference latencies with batch size 1 on a Tesla A100 GPU and a single-core CPU. Our models provide high-quality embeddings on a low-compute budget.}

    \label{fig:result_inference_latency}
\end{figure}

Finally, in Figure~\ref{fig:result_inference_latency} we show that our models
not only provide the best representations, but also offers substantially lower
inference latencies on both high-end GPUs and a single-core CPU. Moreover, training CharEmb Large on an embedding matrix with a vocabulary of five
million entries takes only \textit{three hours} on a single NVIDIA Tesla A100, which
is a negligible time compared to the 10 days required to train SimCSE on the same dataset.

\section{Related Work}

Static embeddings trained with self-supervised setups became popular with
word2vec~\cite{word2vec} and GloVe~\cite{Glove}. While those algorithms have
been originally introduced to embed individual tokens, the approach can be
generalized to entire phrases or multiple token entities by preprocessing
training corpora such that multiple tokens are merged into one.
FastText~\cite{bojanowski2016enriching} can be seen as an extension to Word2Vec
which relies on n-grams to extract representations of unseen text.

Contextualized embeddings (e.g., Elmo~\cite{peters-etal-2018-deep}) are
created by taking into account the contex of each token. Sentence
encoders~\cite{schuster-etal-2019-cross, cer-etal-2018-universal} are a
generalization of contextual embeddings. They can be trained on sentence-level
tasks using supervised datasets, such as NLI, or with unsupervised methods
~\cite{gao-etal-2021-simcse,wang-etal-2021-tsdae-using}.~Our~method to train
text encoders is fully unsupervised and provides higher-quality
representations than supervised encoders.

Embedding compression is a topic of great interest not only for natural
language processing
~\cite{DBLP:conf/mlsys/PansareKACSTV22,Liao_Chen_Wang_Qiu_Yuan_2020}, but also
in recommender systems~\cite{10.1145/3383313.3412227,
kim-etal-2020-adaptive,10.1145/3394486.3403059}. The primary goal of our work
is not to reduce the size of a static embedding matrix, but rather to
generalize the embeddings to entries not seen at training time.

Work has been done to align embedding spaces coming from different models
\cite{joulin-etal-2018-loss,schuster-etal-2019-cross, pmlr-v89-grave19a}.
Instead of aligning spaces coming from static embeddings and sentence encoders,
we introduce text encoders trained to project text in the same space of a
static embedding matrix used as a training dataset.

\section{Conclusion}

Creating embeddings for terms and phrases is of paramount importance for
complex natural language processing platforms targeting highly-technical
domains. While out-of-the-box pre-trained sentence encoders are often
considered as baselines, representations of \textit{similar quality} can be
obtained with substantially lighter and simpler character-based models which are
\textit{5 times smaller} in size and \textit{10 times faster} at
inference time, even on high-end GPUs. The key to obtaining such results is to
realize that large static embedding matrices storing representations for tokens
and terms constitute a very rich supervised dataset to train text encoders
working at the character level. Since both term extraction and embedding
training can be performed without any labeled data, we have proposed a method
to train text encoders which does not require any label. Those models are trained with
the objective of reconstructing the original embedding matrix and can not only
be used as lighter alternatives to sentence encoders, but also as lossless
compressors for large embedding~matrices.

\bibliographystyle{acl_natbib}


\appendix

\begin{table*}[!t]
\centering
    \begin{tabular}{
    @{}l@{\hspace{2mm}}c@{\hspace{2mm}}c@{\hspace{2mm}}c@{\hspace{2mm}}c@{\hspace{2mm}}c@{\hspace{2mm}}c@{\hspace{2mm}}c@{\hspace{2mm}}c@{\hspace{2mm}}c@{\hspace{2mm}}c@{\hspace{2mm}}c@{}}
         & & & & \multicolumn{8}{c}{Correlation}\\
         \cmidrule{5-12}
         & & & & \multicolumn{2}{c}{Original} & & \multicolumn{2}{c}{Reconstr.} & & \multicolumn{2}{c}{Context.}\\
         \cmidrule{5-6}\cmidrule{8-9}\cmidrule{11-12}
        Pre-trained embedding & |Voc.| & Size (MB) & Dim. & Pear. & Spear. & & Pear. & Spear. & & Pear. & Spear.\\
\toprule
GloVe (6B) & 0.4M & 458 & 300 & $42.37$ & $43.95$ & & $36.82	$ & $41.15$ & & $27.36$	& $31.36$\\
GloVe (42B) & 1.9M & 2,194& 300 & $40.30$ & $45.83$ & & $29.89$ & $42.93$ & & $20.66$ & $21.35$\\
GloVe (840B) & 2.2M & 2,513& 300 & $44.83$ & $49.71$ & & $39.32$ & $47.39$ & & $18.97$ & $22.79$\\
FastText wiki-news (16B)& \multirow{1}{*}{1.0M} & \multirow{1}{*}{1,144} & \multirow{1}{*}{300} & $39.01$ & $46.03$ & & $30.72	$ & $45.66$ & & $28.82$ & $27.47$\\
FastText crawl (600B)& \multirow{1}{*}{2.0M} & \multirow{1}{*}{2,289} & \multirow{1}{*}{300} & $47.36$ & $49.32$ & & $45.91	$ & $49.79$ & & $34.49$ & $35.67$\\
Word2Vec news (100B) & 3.0M & 2,861 & 250 & $44.04$ & $44.72$ & & $40.77	$ & $45.28$ & & $45.54$ & $46.09$\\
ArXiv-HUPD uni (2.2B)& \multirow{1}{*}{1.8M} & \multirow{1}{*}{1,403} & 200 & $50.82$ & $52.97$ & & $54.28$ & $55.21$ & & $47.58$ & $45.6$\\
\midrule
ArXiv-HUPD uni + terms (1.8B)& 5.2M & 3,984 & 200 & $\textbf{51.62}$ & $\textbf{53.91}$ & & $\textbf{55.97}$ & $\textbf{57.27}$ & & $\textbf{59.84}$& $\textbf{60.52}$\\	

    \end{tabular}
    \caption{\label{app:tab:all}Additional results when training \textit{CharEmb Large (86MB)} on standard pre-trained word embedding matrices. Without multiword expression, the reconstruction and contextualized via prediction performance are limited.}
\end{table*}

\section{Training Details}
\label{app:training}

To perform the experiments described in Table~\ref{tab:context_dep_emb} we use pre-trained
models publicly available in HuggingFace:

\medskip
\noindent
\textbullet \ \textbf{BERT}:

\href{https://huggingface.co/bert-large-uncased}{bert-large-uncased}.

\medskip
\noindent
\textbullet \ \textbf{Patent-BERT}:

\href{https://huggingface.co/anferico/bert-for-patents}{anferico/bert-for-patents}.

\medskip
\noindent
\textbullet \ \textbf{Sentence-BERT}:

\href{https://huggingface.co/sentence-transformers/all-mpnet-base-v2}{sentence-transformers/all-mpnet-base-v2}. 

\medskip
\noindent
\textbullet \ \textbf{Supervised-SimCSE}:

\href{https://huggingface.co/princeton-nlp/sup-simcse-bert-base-uncased}{princeton-nlp/sup-simcse-bert-base-uncased}.

\medskip
\noindent
\textbullet \ \textbf{Unsupervised-SimCSE}:

\href{https://huggingface.co/princeton-nlp/unsup-simcse-bert-base-uncased}{princeton-nlp/unsup-simcse-bert-base-uncased}.

\medskip
Regarding hyperparameter tuning, the baselines do not need any tuning since all experiments are in a zero-shot fashion. For EmbChar, we only tune the encoder by using LSTM, GRU, or Transformer on the validation set. More specifically, we split the word embedding matrix into train and validation sets with a ratio of 80-20. No other hyperparameters have been explored. We stop the training of EmbChar using early-stopping on the validation set when the average cosine similarity has not been improved since 10 epochs. 

Our hyperparameters are shown in Table~\ref{tab:app_params}.
We train our word embedding with Word2Vec with CBOW and a window size of 8 and 25 epochs. For FastText, we kept the default parameters.

All experiments have been run on the following hardware:

\medskip
\noindent
\textbullet \ \textbf{CPU}: AMD EPYC 7763 64-core processor.

\medskip
\noindent
\textbullet \ \textbf{RAM}: 1.96~TB.

\medskip
\noindent
\textbullet \ \textbf{GPU}: NVIDIA Tesla A100.

\medskip
\noindent
\textbullet \ \textbf{OS}: Red Hat Enterprise Linux 8.6.

\medskip
\noindent
\textbullet \ \textbf{Software}: PyTorch 1.12.1, CUDA 11.6.

\medskip
We emphasise that we train the word embedding matrices on a 16-core virtual
machine hosted on \textit{AMD EPYC 7742}. An epoch takes approximately 25
minutes. Training our embedding matrix \textit{ArXiv-HUPD uni + terms} requires
\textit{less} than 10 dollars of compute budget in the cloud. Training our
model EmbChar Large thereafter takes a few hours on a single NVIDIA Tesla A100,
costing approximately \$5 to \$10\footnote{The hourly pricing for spot instances with one
A100 GPU is in the range \$1.25-\$1.5 in public cloud offerings.}. In contrast, training
SimCSE on the same dataset takes around 10 days.

\begin{table}[!h]
\small
\centering
    \begin{tabular}{
    @{}lcccc@{}}
    & EmbChar & EmbChar & EmbChar\\
    Hyperparameter & Small & Base & Large\\
\toprule
Hidden dimension & 512 & 512 & 768\\
Number of layers & 1 & 2 & 2\\
Bidirectional & True & True & True\\
Dropout & 0.2 & 0.2 & 0.2\\
Learning rate & 0.001 & 0.001 & 0.0005\\
Weight decay & 1e-8 & 1e-8 & 1e-8\\
Batch size & 256 & 256 & 256\\
    \end{tabular}
    \caption{\label{tab:app_params}The hyperparameter for all EmbChar variants.}
\end{table}

\section{Additional Results}
\label{app:context}

In Table~\ref{app:tab:all} we report the results for the experiments in
Section~\ref{sec:evaluation} when training our CharEmb Large (86MB) on
different word embedding matrices. For each of the embedding matrix considered,
we measure the Pearson and Spearman correlation for the three setups: i) the
original embedding matrix~(\textit{Original}), ii) an embedding matrix
reconstructed using the same vocabulary of the original
one~(\textit{Reconstr.}), and iii) when the embedding of given terms are
contextual, i.e., predicted with a CharEmb model trained over the original
matrix~(\textit{Context.}). The table showcases multiple important findings.
First, among the pre-trained models, the vocabulary size plays a significant
role to achieve high correlation, with the Word2Vec model with a vocabulary of
3 million entries outperforming embedding matrices that have been trained over
larger datasets ~(e.g., Glove 840B or FastText crawl). Second, the domain of
the corpus used to train the embeddings plays a significant role. By training
with a corpus of only 2 billion in-domain tokens, an embedding matrix with a
vocabulary of 1.8 million entries achieves similar correlation of much
larger embedding matrices. Third, our CharEmb model achieves the best
performance when trained with an embedding matrix containing embeddings for
terms. Predicting the embeddings with our CharEmb model allows to achieve
significantly higher correlation than the original matrix containing terms.

\end{document}